\pdfoutput=1

\documentclass[11pt]{article}
\usepackage[]{eacl2024}

\usepackage{times}
\usepackage{latexsym}

\usepackage[T1]{fontenc}

\usepackage[utf8]{inputenc}

\usepackage{microtype}

\usepackage{inconsolata}

\usepackage{enumitem}
\usepackage{relsize}
\usepackage{booktabs}
\usepackage{tabularx}
\usepackage{multirow}
\usepackage{graphicx}
\usepackage{xspace}
\usepackage{xcolor}
\usepackage{caption}
\usepackage{subcaption}
\usepackage{listings}

\usepackage{amsmath}
\usepackage{amssymb}
\usepackage{amsthm}
\usepackage{amsfonts}
\usepackage{mathtools}

\usepackage{tikz}
\usetikzlibrary{arrows.meta}
\usetikzlibrary{shapes.arrows}
\usetikzlibrary{shapes.misc}
\usetikzlibrary{fit}
\usetikzlibrary{calc,shapes}
\usetikzlibrary{tikzmark}
\usetikzlibrary{arrows}
\usetikzlibrary{trees}

\usepackage{cleveref}

\usepackage{placeins}
\usepackage{float}
\crefname{chapter}{Chapter}{}
\crefname{section}{Section}{Sections}
\crefformat{section}{Section~#2#1#3}

\crefname{table}{Table}{Tables}
\crefname{figure}{Figure}{Figures}
\crefname{algorithm}{Alg.}{Algs.}
\crefname{line}{Line}{Lines}
\crefname{appendix}{App.}{}
\crefname{chapter}{Chapter}{Chapters}

\crefname{thm}{Theorem}{Theorems}
\crefname{prop}{Proposition}{Propositions}
\crefname{definition}{Definition}{Definitions}
\crefname{lemma}{Lemma}{Lemmas}
\crefname{cor}{Corollary}{Corollaries}
\crefname{equation}{Eq.}{Eqs.}

\creflabelformat{equation}{#2\textup{#1}#3}
\crefrangelabelformat{equation}{(#3#1#4--#5#2#6)}

\theoremstyle{definition}

\newcolumntype{C}{>{\centering\arraybackslash}X}

\renewcommand{\ldots}{\ensuremath{{\ldotp\kern-0.2em\ldotp\kern-0.2em\ldotp}}}

\renewcommand{\cdots}{\ensuremath{{\cdotp\kern-0.2em\cdotp\kern-0.2em\cdotp}}}
\renewcommand{\dots}{\ensuremath{{\ldotp\kern-0.2em\ldotp\kern-0.2em\ldotp}}}

\newcommand{\defn}[1]{\textbf{#1}}

\newcommand{\defeq}[0]{\mathrel{\stackrel{\textnormal{\tiny def}}{=}}}

\newcommand{\checkNotation}[1]{{#1}} 

\newcommand{\checkNotationOp}[3]{\checkNotation{#1 {\normalcolor {#2}} #3}}

\newcommand{\abs}[1]{\checkNotationOp{\left\lvert}{#1}{\right\rvert}}

\newcommand{\json}{\checkNotation{JSON}\xspace}

\newcommand{\predTree}{\checkNotation{T}}
\newcommand{\trueTree}{\checkNotation{G}}
\newcommand{\precision}{\checkNotation{\mathrm{prec}}}
\newcommand{\recall}{\checkNotation{\mathrm{rec}}}

\usepackage[plain]{algorithm}
\usepackage{algorithmicx}
\usepackage[noend]{algpseudocode}
\algnewcommand{\parState}[1]{\State%
    \parbox[t]{\dimexpr\linewidth-\algmargin}{\strut\hangindent=\algorithmicindent \hangafter=1 #1\strut}}
\algrenewcommand\algorithmicindent{1.0em}%
\newcommand{\algorithmicdowhile}{\textbf{do}:}
\algdef{SE}[DOWHILE]{Do}{doWhile}{\algorithmicdowhile}[1]{\algorithmicwhile\ #1}%
\newcommand{\algorithmicfunc}[1]{\textbf{def} #1 :}
\algdef{SE}[FUNC]{Func}{EndFunc}[1]{\algorithmicfunc{#1}}{}

\newcommand{\algorithmicclass}[1]{\textbf{class} #1 :}
\algdef{SE}[CLASS]{Class}{EndClass}[1]{\algorithmicclass{#1}}{}

\newif\ifboldnumber

\algrenewcommand\alglinenumber[1]{%
  \footnotesize\ifboldnumber\color{red}\bfseries\fi\global\boldnumberfalse#1:}

\MakeRobust{\Call}   

\newcommand{\rightcomment}[1]{{\color{commentcolor} \(\triangleright\) {\footnotesize\textit{#1}}}}
\algrenewcommand{\algorithmiccomment}[1]{\hfill \rightcomment{#1}}  
\algnewcommand{\LineComment}[1]{\State \rightcomment{#1}}
\algnewcommand{\LinesComment}[1]{\State \rightcomment{\parbox[t]{\linewidth-\leftmargin-\widthof{\(\triangleright\) }}{#1}}}

\renewcommand\algorithmicthen{:}

\makeatletter
\ifthenelse{\equal{\ALG@noend}{t}}%
  {\algtext*{EndFunc}}
  {}%
\makeatother

\makeatletter
\ifthenelse{\equal{\ALG@noend}{t}}%
  {\algtext*{EndClass}}
  {}%
\makeatother

\algnewcommand{\IIf}[1]{\State\algorithmicif\ #1\ \algorithmicthen}
\algnewcommand{\EndIIf}{\unskip}

\definecolor{commentcolor}{rgb}{0.4, 0.22, 0.33}

\colorlet{punct}{red!60!black}
\definecolor{background}{HTML}{EEEEEE}
\definecolor{delim}{RGB}{20,105,176}
\colorlet{numb}{magenta!60!black}

\lstdefinelanguage{json}{
    basicstyle=\footnotesize\ttfamily,
    numbers=none,
    numberstyle=\scriptsize,
    stepnumber=1,
    numbersep=8pt,
    showstringspaces=false,
    breaklines=true,
    literate=
      {:}{{{\color{punct}{:}}}}{1}
      {,}{{{\color{punct}{,}}}}{1}
      {\{}{{{\color{delim}{\smaller \{}}}}{1}
      {\}}{{{\color{delim}{\smaller \}}}}}{1}
      {[}{{{\color{delim}{[}}}}{1}
      {]}{{{\color{delim}{]}}}}{1},
}

\title{TreeForm: End-to-end Annotation and Evaluation for Form Document Parsing}

\author{Ran Zmigrod,~\;~Zhiqiang Ma,~\;~Armineh Nourbakhsh,~\;~Sameena Shah \\
  J.P. Morgan AI Research \\
  \texttt{\{first\_name\}.\{last\_name\}@jpmchase.com}
  }

\allowdisplaybreaks

\renewcommand{\paragraph}[1]{\vspace{5pt}\noindent\textbf{#1.}}

\begin{document}
\maketitle
\begin{abstract}
Visually Rich Form Understanding (VRFU) poses a complex research problem
due to the documents' highly structured nature and yet highly variable style and content.
Current annotation schemes decompose form understanding and omit key hierarchical structure, making development and evaluation of end-to-end models difficult.
In this paper, we propose a novel F1 metric to evaluate form parsers and describe a new content-agnostic, tree-based annotation scheme for VRFU: \defn{TreeForm}.
We provide methods to convert previous annotation
schemes into TreeForm structures and evaluate TreeForm predictions using a modified version of the normalized tree-edit distance.
We present initial baselines for our end-to-end performance metric and the TreeForm edit distance, averaged over the FUNSD and XFUND datasets, of $61.5$ and $26.4$ respectively. 
We hope that TreeForm encourages deeper research in  annotating, modeling, and evaluating the complexities of form-like documents.
\end{abstract}

\section{Introduction}
Visually rich document understanding (VRDU) has been a growing field in multimodal AI research.
VRDU takes document images as input and applies tasks such as document classification \citep{gu-21-unidoc, kim-22-ocr, gu-22-unified}, information extraction \citep{borchmann-21-due, wang-22-mmlayout}, visual question-answering (VQA) \citep{mathew-21-docvqa, tito-21-icdar, li-22-text}, \emph{inter alia}.
Consequently, state-of-the-art VRDU models rely on image-to-text understanding, such as optical character recognition (OCR), as well as multimodal models that can exploit textual, visual, and spatial features of documents.

VRFU is a subset of VRDU that focuses on forms, 
i.e., documents that contain a collection of hierarchical key-value pairs (in various styles) regarding specific entities.
Forms are highly structured documents, and so can be directly parsed into a structured object. 
Parsing receipts, which share a similar though smaller and simpler structure, have been the focus of much document parsing research \citep{majumder-20-representation, borchmann-21-due, gao-21-field}.
In this work, we focus on the broader field of form parsing.

\definecolor{FormBurgundy}{RGB}{121,26,62}
\newcommand{\burgundy}{{\color{FormBurgundy} burgundy\xspace}}
\definecolor{FormGreen}{RGB}{79,122,40}
\newcommand{\green}{{\color{FormGreen} green\xspace}}
\definecolor{FormBlue}{RGB}{0,86,214}
\newcommand{\blue}{{\color{FormBlue} blue\xspace}}
\definecolor{FormOrange}{RGB}{255,106,0}
\newcommand{\orange}{{\color{FormOrange} orange\xspace}}
\definecolor{FormPink}{RGB}{211,87,254}
\newcommand{\pink}{{\color{FormPink} pink\xspace}}

\begin{figure}
    \centering
    \includegraphics[width=\linewidth]{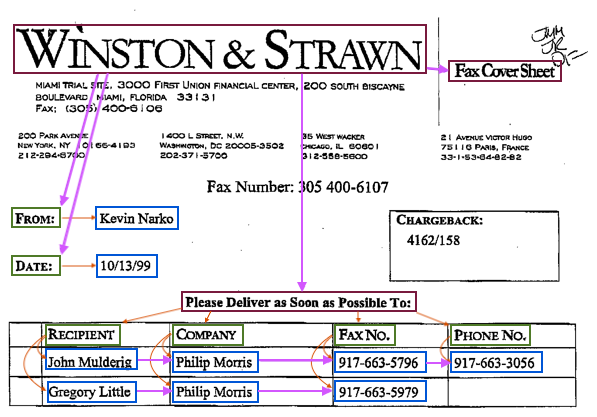}
    \caption{Excerpt of a FUNSD form. Headers are marked in \textbf{\burgundy{}}, questions are marked \textbf{\green{}}, and answers are marked in \textbf{\blue{}}. Entity links provided by the FUNSD annotation schemes are marked in \textbf{\orange{}}. Links in \textbf{\pink{}} were created for TreeForm.}
    \label{fig:funsd-example}
\end{figure}

The FUNSD dataset \citep{jaume-19-funsd}, and its multilingual counter-part XFUND \citep{xu-22-xfund}, are standard VRFU datasets used in the literature.
These datasets split form parsing into entity labeling and entity linking tasks.
Unfortunately, this annotation scheme is unable to fully express the structure of a form.
Moreover, to the best of our knowledge, no holistic approach exists for evaluating complete FUNSD-type predictions.
Few other form parsing datasets exist, such as FUNSD+ \citep{zagami-22-funsd}
and the National Archives Forms (NAF) dataset \citep{davis-19-deep}.
We do not address these directly but many of our contributions can be extended to these annotation schemes.

In this paper, we tackle the problem of complete form parsing through a tree-based approach that successfully captures the hierarchical structure of forms which is often missed in modern annotation schemes. Our contributions are summarized below:
\begin{enumerate}[noitemsep]
    \item We provide novel end-to-end metrics for evaluating FUNSD-type annotations.
    \item We present a novel tree-based representation of forms, \defn{TreeForm}, that is conveniently stored as a single \json object.
    We also utilize the greedy-aligned tree-edit distance (GAnTED) proposed by \citet{davis-22-end} to evaluate TreeForm predictions.
    \item We detail a method to transform \emph{any} FUNSD-type annotations into TreeForm. Our method captures additional hierarchical and tabular information, which FUNSD annotations do not contain.
    The additional information gained from TreeForm is visualized in \cref{fig:funsd-example}.\footnote{Note that for ease of visualization, \cref{fig:funsd-example} does not contain boxes for every single annotation. In practice, every piece of information annotated in FUNSD (or the original annotation scheme) will be captured in TreeForm.}
    \item We evaluate TreeForm and our novel metrics using state-of-the-art models \citep{kim-22-ocr, xu-22-xfund} on the FUNSD and XFUND datasets. Our baselines achieve a node-alignment accuracy of $0.22$, end-to-end F1 score of $61.5$, and TreeForm GAnTED score of $14.5$ across all languages.
\end{enumerate}

\section{Limitations of Form Parsing Datasets}\label{sec:funsd}
Most VRFU research employs FUNSD-type annotations \citep{jaume-19-funsd}.
This annotation scheme contains two components: First, a list of semantic entities defined by a group of tokens and a semantic label (header, question, or answer).\footnote{The original annotations contain a fourth entity, \emph{other}, for text that does not conform to one of the three aforementioned labels.
We follow recent work and ignore these labels \citep{xu-21-layout, xu-21-layoutlm2, xu-22-xfund}.}
Second, a list of directed links between entities to denote form structure.
An example of an FUNSD annotation is given in \cref{app:json}.
While the FUNSD (and XFUND) project was a fundamental step in VRFU, its approach contains underlying issues \citep{vu-20-revising}.
Other non-FUNSD schematic VRFU datasets exist \citep{davis-19-deep, zagami-22-funsd}.
However, they are not widely used and are not without similar issues.
\cref{tab:data} provides an overview of form parsing datasets.

\begin{table}[t]
    \centering
    \small
    \begin{tabular}{l l p{1.5cm} c c}
         \bf Dataset & \bf Lang. & \bf Scheme & \bf $\mid$Train$\mid$ & \bf $\mid$Test$\mid$ \\ \midrule
         FUNSD 
         & EN & FUNSD & $149$ & $50$ \\
         XFUND 
         & Multi & FUNSD & $1043$ & $350$ \\
         FUNSD+ 
         & EN & FUNSD+ & $1023$ & $116$ \\ 
         NAF 
         & EN & NAF & $741$
         & $63$
    \end{tabular}
    \caption{Form Understanding Datasets. XFUND is equally split over seven languages ($149$ training documents, $50$ test documents): DE, ES, FR, IT, JA, PT, ZH.}
    \label{tab:data}
\end{table}

\paragraph{Task Decomposition}
The FUNSD annotation scheme proposes that VRFU should be decomposed into semantic entity labeling and entity linking.
This makes a proper end-to-end evaluation of form understanding models difficult as combining the two task metrics does not represent an accurate joint evaluation.
Indeed, works that report individual F1 scores do not attempt to combine these \citep{carbonell-20-named, li-21-structext, gemelli-22-doc, hong-22-bros} while much research omits the entity linking task \citep{li-20-structural, appalaraju-21-docformer, xu-21-layout, xu-21-layoutlm2, chen-22-xdoc, luo-22-bi}.

\paragraph{Inconsistent Annotations}
The FUNSD annotations contain several inconsistencies in both their entity labeling and entity linking components.
Inconsistencies arise in hierarchical forms, handwritten input, and missing entity links.
While many of these inconsistencies were improved in \citep{vu-20-revising}, some issues still persist 

\paragraph{Table Recognition}
Tables are a natural component of forms, however, they break the common question-answer structure.
The FUNSD annotation scheme handles tables as columns, where a column header is a question and each value in the column is an answer linked to the question.
Row headers are sometimes also annotated and linked, but this is a less common pattern in the dataset.
Also, as pointed out by \citet{davis-22-end}, row and column alignment is not present in the annotations, making table parsing or reconstruction not possible in FUNSD.

\section{An End-to-end Evaluation of FUNSD}\label{sec:end-to-end}
\begin{figure*}[!ht]
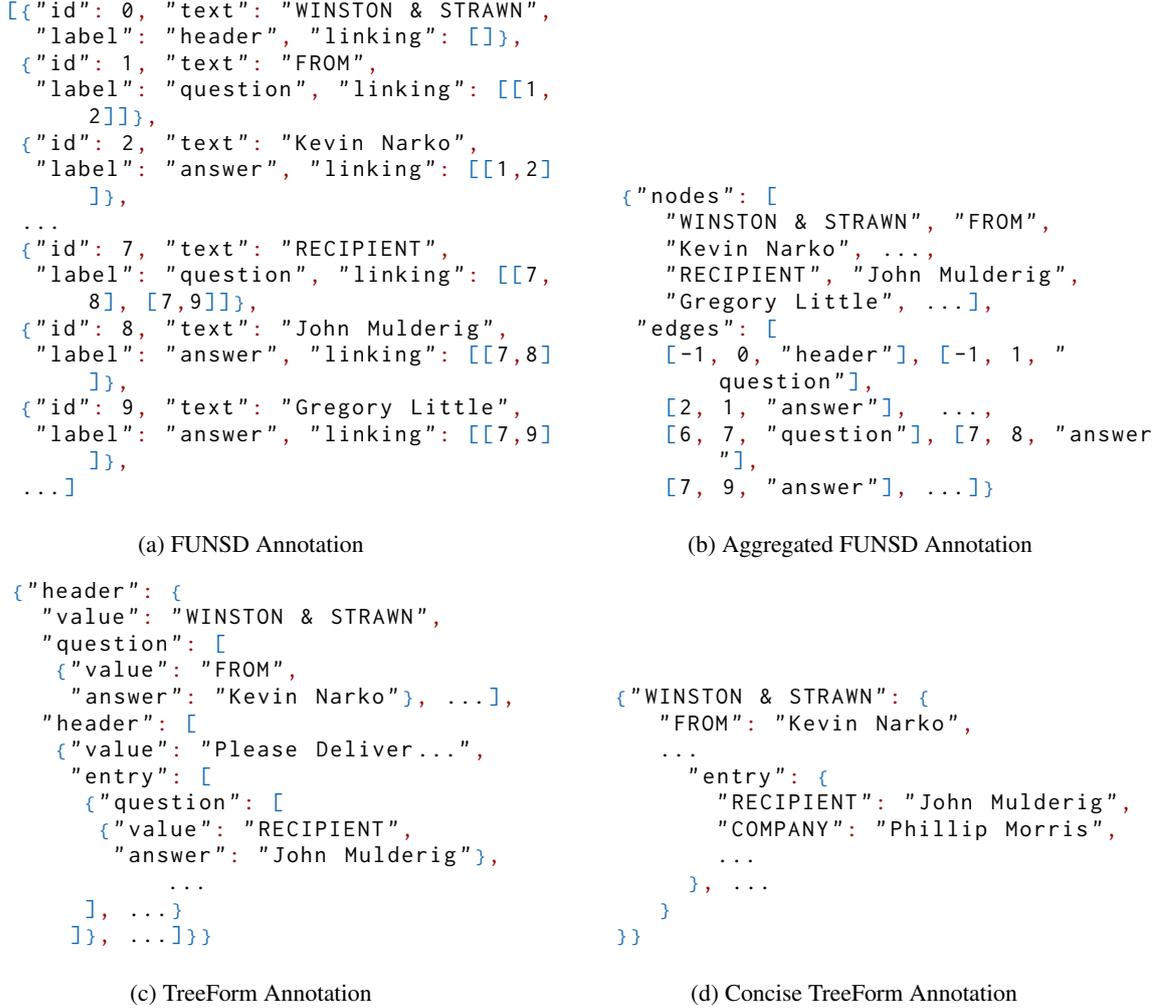

    \centering
    \begin{subfigure}[b]{0.5\linewidth}
    \begin{lstlisting}[basicstyle=\tiny,language=json,firstnumber=1,]
    [{"id": 0, "text": "WINSTON & STRAWN",
      "label": "header", "linking": []},
     {"id": 1, "text": "FROM",
      "label": "question", "linking": [[1,2]]},
     {"id": 2, "text": "Kevin Narko",
      "label": "answer", "linking": [[1,2]]},
     ...
     {"id": 7, "text": "RECIPIENT",
      "label": "question", "linking": [[7,8], [7,9]]},
     {"id": 8, "text": "John Mulderig",
      "label": "answer", "linking": [[7,8]]},
     {"id": 9, "text": "Gregory Little",
      "label": "answer", "linking": [[7,9]]},
     ...]
    \end{lstlisting}
    \caption{FUNSD Annotation}
    \label{fig:example:funsd}
    \end{subfigure}
    \begin{subfigure}[b]{0.49\linewidth}
    \begin{lstlisting}[language=json,firstnumber=1]
    {"nodes": [
       "WINSTON & STRAWN", "FROM",
       "Kevin Narko", ...,
       "RECIPIENT", "John Mulderig",
       "Gregory Little", ...],
     "edges": [
       [-1, 0, "header"], [-1, 1, "question"],
       [2, 1, "answer"],  ...,
       [6, 7, "question"], [7, 8, "answer"],
       [7, 9, "answer"], ...]}
    \end{lstlisting}
    \caption{Aggregated FUNSD Annotation}
    \label{fig:example:end-to-end}
    \end{subfigure}

     \begin{subfigure}[b]{0.49\linewidth}
    \begin{lstlisting}[language=json,firstnumber=1]
    {"header": {
      "value": "WINSTON & STRAWN",
      "question": [
       {"value": "FROM",
        "answer": "Kevin Narko"}, ...],
      "header": [
       {"value": "Please Deliver...",
        "entry": [
         {"question": [
          {"value": "RECIPIENT",
           "answer": "John Mulderig"}, ...
         ], ...}
        ]}, ...]}}
    \end{lstlisting}
    \caption{TreeForm Annotation}
    \label{fig:example:treeform}
    \end{subfigure}
    \begin{subfigure}[b]{0.49\linewidth}
    \begin{lstlisting}[language=json,firstnumber=1]
    {"WINSTON & STRAWN": {
       "FROM": "Kevin Narko",
       ...
         "entry": {
           "RECIPIENT": "John Mulderig",
           "COMPANY": "Phillip Morris", 
           ...
         }, ...
       }
    }}
    \end{lstlisting}
    \caption{Concise TreeForm Annotation}
    \label{fig:example:concise-treeform}
    \end{subfigure}
    \caption{Different annotation schemes for excerpt of FUNSD form given in \cref{fig:funsd-example}.}
    \label{fig:example}
\end{figure*}

In this section, we propose a new F1 metric to evaluate FUNSD-type predictions on their combined entity labeling and entity linking performance.
Our metric, inspired by the labeled attachment score (LAS) \citep{2009-kubler-dependency}, is based on constructing a tree of the FUNSD annotations.
Each node in the tree represents a semantic FUNSD entity, and each edge in the tree is a link between entities, labeled with the entity label of the child node.
Entities with no incoming links have an incoming edge from a dummy root node, labeled with the entity's label.
An example of this structure is given in \cref{fig:example:end-to-end}.
Let $\predTree$ be the set of predicted edges and $\trueTree$ be the set of ground truth edges.
We can then respectively define the precision and recall as:\looseness=-1
\begin{align*}
    \precision(\predTree, \trueTree) \defeq \frac{\abs{\predTree \cap \trueTree}}{\abs{\predTree}} \quad    \recall(\predTree, \trueTree) \defeq \frac{\abs{\predTree \cap \trueTree}}{\abs{\trueTree}}
\end{align*}
These can be combined as normal to yield an F1 metric that examines both entity labels and links.
Note that this metric equally punishes incorrect edges regardless of where they lie in the nested tree.
For example, mis-predicting a top level node (i.e., one connected to the dummy root node) is punished equally to mis-predicting the edge connecting an answer to a question.

The above F1 metric does not consider node-alignment (i.e., word grouping) in its evaluation.
Indeed, it assumes that the nodes of the predicted and ground truth trees are aligned.
Node alignment can be done through a greedy algorithm.
For each predicted entity, we compute the normalized Levenshtein distance \citep{mathew-21-docvqa} to each true entity.\footnote{We choose the Levenshtein distance to follow related work and to enable measuring the edit distance in a more meaningful way as we can assign context-specific penalties to additions, deletions, and alterations.}
We then greedily select alignments that meet a certain threshold (edit distance less than $0.4$), until no possible alignments remain.
We define the node-alignment accuracy (NAA) metric as the mean normalized Levenshtein distance between each pair of aligned nodes.\footnote{For unaligned nodes, we use the normalized Levenshtein distance between a node and the empty string, i.e., $1$.}\looseness=-1

\section{A Tree-based Annotation Scheme}\label{sec:treeform}
In this section we present a novel tree-based form annotation scheme, \defn{TreeForm}.
Unlike our F1 and NAA metrics, TreeForm does not assume the FUNSD task decomposition.
Instead, TreeForm provides a fully structured object that enables complete representation of a form.

A TreeForm node is represented by a \json object (i.e., a dictionary) whose key and value respectively contain the node's (textual) value and a dictionary where each key is either another node or an edge label that leads to a group of child nodes.
Therefore, a TreeForm tree is represented using a single \json object which efficiently contains all form information necessary for any down-stream VRFU task.
\citet{davis-22-end} also parses complete forms into \json objects, however, their approach requires multiple objects per form and so annotations are less readable.
TreeForm considers forms to be comprised of three key structures: headers, question-answer pairs, and tables; each of these structures is represented in a tree structure which is detailed in \cref{app:treeform}.
An example of a TreeForm annotation is given in \cref{fig:example:treeform}.

\begin{table*}[t]
    \centering
    \small
    \begin{tabular}{l l c c c c c c c c}
         \multirow{2}{*}{\bf Metric} & \multirow{2}{*}{\bf Model} & \bf FUNSD & \multicolumn{7}{c}{\bf XFUND}\\
         & & \bf EN & \bf DE & \bf ES & \bf FR & \bf IT & \bf JA & \bf PT & \bf ZH \\ \midrule
         Labeling F1 $\uparrow$ & \multirow{2}{*}{LayoutXLM} &
         $77.6$ & $79.5$ & $74.1$ & $78.9$ & $77.3$ & $79.1$ & $79.4$ & $88.9$ \\
         Linking F1 $\uparrow$ & &
         $48.8$ & $57.3$ & $63.8$ & $61.4$ & $53.5$ & $64.4$ & $53.0$ & $67.8$
         \\ \midrule
         \multirow{2}{*}{NAA $\downarrow$} 
         & LayoutXLM & $\mathbf{0.21}$ & $\mathbf{0.24}$ & $\mathbf{0.28}$ & $\mathbf{0.23}$ & $\mathbf{0.23}$ & $\mathbf{0.24}$ & $\mathbf{0.23}$ & $\mathbf{0.12}$ \\ 
         & Donut & $0.69$ & $0.58$ & $0.69$ & $0.61$ & $0.64$ & $0.75$ & $0.75$ & $0.74$ \\ 
         \midrule
         \multirow{2}{*}{Tree F1 $\uparrow$}
         & LayoutXLM & $62.6$ & $57.0$ & $\mathbf{60.6}$ & $\mathbf{65.7}$ & $59.6$ & $\mathbf{59.3}$ & $\mathbf{59.9}$ & $\mathbf{66.9}$ \\ 
         & Donut & $\mathbf{64.8}$ & $\mathbf{61.5}$ & $53.0$ & $62.4$ & $\mathbf{62.9}$ & $51.7$ & $46.8$ & $55.6$ \\ 
         \midrule
         \multirow{2}{*}{GAnTED $\downarrow$}
         & LayoutXLM & $18.0$ & $27.7$ & $20.7$ & $22.3$ & $20.6$ & $7.2$ & $25.1$ & $\mathbf{7.3}$ 
         \\ 
         & Donut & $\mathbf{15.1}$ & $\mathbf{21.0}$ & $\mathbf{15.8}$ & $\mathbf{13.6}$ & $\mathbf{17.5}$ & $\mathbf{6.7}$ & $\mathbf{17.3}$ & $8.9$ \\
    \end{tabular}
    \caption{Standard and end-to-end evaluation metrics for FUNSD and TreeForm annotations.
    }
    \label{tab:experiments-funsd}
\end{table*}

\subsection{Evaluating TreeForm}
We assess TreeForm predictions using a variant of the normalized tree-edit distance (nTED) \citep{hwang-21-spatial}, a distance metric that informs us how much work is needed to convert the predicted tree into the ground-truth tree (and so a lower nTED is better).
Specifically, we follow the greedy-aligned nTED (GAnTED) metric proposed by \citet{davis-22-end}.
This is a more holistic metric than those proposed in \cref{sec:end-to-end} as it does not assume any task decomposition.

\subsection{Converting FUNSD into TreeForm}\label{sec:treeform:transform}
We propose a set of transformations and heuristics to convert FUNSD-type annotations into TreeForm structures.
We first apply simple conversions for straight-forward cases such as single-answer questions and hierarchical headers which fit directly into the TreeForm annotation scheme.
Next, we elaborate on the more complex heuristics we have chosen to create the most accurate and complete TreeForm structure.
Note an important caveat to our transformations, TreeForm cannot attain information that was not annotated in FUNSD (or the original annotation scheme).

\paragraph{Discarding Incomplete Information}
TreeForm does not include free-form text, unanswered questions, or unprompted answers.
Any such annotations
in the dataset are discarded here.

\paragraph{Choosing the Form Title}
If headers exist in the form, we aim to assign a form title to each TreeForm structure.
Since FUNSD does not necessary nest headers correctly (as seen in \cref{fig:funsd-example}), we consider the form title to be the non-nested header that is spatially highest on the page.
This seems like a sensible heuristic as we typically expect titles to be at the top of documents.\footnote{
This heuristic is not perfect, as for example, the form title in \cref{fig:funsd-example} could be ``Fax Cover Sheet'' rather than ``WINSTON \& STRAWN''; this also raises an inconsistency issue as it is unclear whether the latter header is indeed a header entity.}

\paragraph{Constructing Tables}
Table structure is not directly recoverable from FUNSD annotations (as discussed in \cref{sec:funsd}).
We utilize bounding box details to align columns and rows such that we can approximate the full table.
Similar heuristics were also used in \cite{davis-22-end}.
Some questions with multiple answers in FUNSD may indicate a multi-line answer rather than a table column (or row).
Therefore, we require multiple answers to start roughly in the same horizontal (or vertical) position to be considered a table entry.

\section{Experiments}\label{sec:experiments}
To evaluate TreeForm as well as our end-to-end FUNSD metrics, we fine-tune LayoutXLM \citep{xu-22-xfund} and Donut \citep{kim-22-ocr} on each of the FUNSD and XFUND datasets.\footnote{We fine-tune one variant of each model on FUNSD annotations and another on TreeForm. Details are given in \cref{app:experiments}.}
We chose these two models as they represent the two current approaches to form parsing: Pipelined\footnote{The pipelined approach first predicts the entity labels and uses said predictions to perform the entity linking task.} (LayoutXLM) and end-to-end (Donut). 
We describe pre- and post-processing steps required for applying the FUNSD and TreeForm evaluations to each model in \cref{app:experiments}.
Importantly, we performed post-processing steps to the output of LayoutXLM to enable evaluation using our metrics.
The results are given in \cref{tab:experiments-funsd}.

For all languages, our end-to-end F1 metric is lower than one or both of the standard labeling and linking F1 scores.
This indicates that while past work may have high labeling (or linking) F1 scores, they may not have full form understanding capabilities.
This is further exacerbated as our F1 metric requires the NAA score to provide a holistic overview.
As expected, LayoutXLM outperforms Donut with respect to NAA as it has access to OCR tokens.
However, Donut still outperforms LayoutXLM on F1 for some languages, suggesting it is better at understanding hierarchical structure.

We consider the median GAnTED score for evaluating TreeForm predictions.
Donut outperformed LayoutXLM in all languages except for Chinese.
This is expected as Donut was designed with document parsing in mind and so could be directly fine-tuned on TreeForm data whereas LayoutXLM was pipelined using the FUNSD annotation scheme (see \cref{app:experiments}).
As GAnTED has not yet been widely used in the literature, it is difficult to assess the effectiveness of these systems.
Nevertheless, Dessurt \cite{davis-22-end} reported a GAnTED score of $23.4$ on the FUNSD dataset which aligns with the scores in \cref{tab:experiments-funsd}.\footnote{
In the experimental set-up of \citet{davis-22-end}, a flatter tree representation (with some other key differences) is used and so the results are not directly comparable.}
Additionally, the GAnTED scores seem to be similar across languages which attests to the generalizability of TreeForm.

\section{Conclusion}
In this paper, we described the challenges of current form understanding and the limitations of the current datasets and their annotation schemes.
We introduced a novel F1 metric that can be applied to current annotation schemes and further proposed a new tree-based annotation scheme, TreeForm, that enables complete form parsing.
We applied state-of-the-art models (LayoutXLM and Donut) to provide the first TreeForm baselines.
We envision future work to create a new TreeForm dataset, that contains correct and consistent annotations.

\section*{Disclaimer}
This paper was prepared for informational purposes by the Artificial Intelligence Research group of JPMorgan Chase \& Co and its affiliates (“JP Morgan”), and is not a product of the Research Department of JP Morgan. JP Morgan makes no representation and warranty whatsoever and disclaims all liability, for the completeness, accuracy or reliability of the information contained herein. This document is not intended as investment research or investment advice, or a recommendation, offer or solicitation for the purchase or sale of any security, financial instrument, financial product or service, or to be used in any way for evaluating the merits of participating in any transaction, and shall not constitute a solicitation under any jurisdiction or to any person, if such solicitation under such jurisdiction or to such person would be unlawful.

\bibliography{eacl2024}
\bibliographystyle{acl_natbib}

\clearpage

\appendix

\section{Related Work}\label{app:related}
We discuss work regarding form information extraction (i.e., entity labeling and entity linking) in the main paper.
Here, we briefly describe other relevant VRFU tasks as well as other domains for document parsing.

\paragraph{Form Structure Extraction}
Complete form parsing is related to form structure extraction, a task which aims to learn the structure and type of field values in forms such as a text field or a checkbox \citep{aggarwal2020multi, aggarwal-etal-2020-form2seq, sarkar-2020-document, gao-21-field, mathur-23-layerdoc}.
The tasks differ in that structure extraction does not aim to extract the answers of a form (indeed the input forms do not need to be filled in), nor does it aim to necessarily learn hierarchical structure.
As such, its datasets do not need to be filled in forms, and can just be form templates.
We note that structure extraction is a possible approach to form parsing. 

\paragraph{Table Detection}
Tables are a natural occurrence in semi-structured documents and so have merited much study in the literature \citep{riba-2019-table, qasim-2019-rethinking, zhong-2019-publay}.
There has been a plethora of work on table understanding that dates back several decades; \citep{zanibbi-2004-survey} provides a through analysis of initial methods for table detection.
More recent work has leveraged image, text, and spatial features to train neural networks to achieve F1 scores of over $80$ for cell and table header detection \citep{schreiber-2017-deep, herzig-2020-tapas, zhong-2020-image, gemelli-2022-graph}.

\paragraph{Receipt and Invoice Parsing}
Document parsing is the task of assigning structure to a document image.
While work exists for a variety of document types such as namecards \citep{hwang-2019-post, hwang-2021-cost} and even forms \citep{davis-22-end}, receipts and invoices have been the most prominent domain for document parsing \citep{huang-2019-comp, majumder-20-representation, borchmann-21-due, gao-21-field, simsa-2023-docile}.
This is largely due to shared characteristics among receipts (e.g., item names and prices) and their structured layout.
Indeed, the end-to-end model used in this work, Donut \citep{kim-22-ocr}, was chosen due to its effectiveness in invoice parsing.

\section{Detailed TreeForm Annotation Scheme}\label{app:treeform}
In this section, we describe the TreeForm construction in more detail.
Specifically we discuss two versions of TreeForm: Concise and non-concise.
We view forms as being comprised of four components: headers, question-answer pairs, tables, and free-form text.\footnote{Question-answer pairs cover any single question response, including checkboxes or similar structures.}
In the non-concise TreeForm, we use tree leaves to  capture all textual information of a form.
Non-leaf nodes are then used to describe the type of textual information (e.g., question, answer) as well as structural information (e.g., question-answer pairs, table entries).
A similar structure is suggested in the supplementary materials of \cite{davis-22-end}.
TreeForm differs in that leaves of the tree always correspond to the textual content of the form.\footnote{We may also want to include bounding box information in TreeForm annotations, this can be done by attaching a child to each text value that contains the node's bounding box.
Note that \cite{davis-22-end} did not offer bounding box information in their proposed tree scheme.
}
We present the non-concise version of TreeForm to show the underlying structure of the form; an example of this annotation scheme is given in \cref{fig:example:treeform}.
We hope that future annotations of TreeForm will not need this more verbose version and will immediately use the structure described in \cref{sec:treeform:concise}.

\paragraph{Headers}
Headers are represented by an intermediate header node with a connected leaf node containing the header value (if a value is given).
A header node contains the sub-tree of all information that falls under that header (and associated section).
Therefore, the header node associated with the form title (if one exists) is the root of the tree.

\paragraph{Question-answer Pairs}
Question-answer pairs are represented by a chained question node and answer node.
The question node has two children: a leaf node containing the question text, and an answer node that has its own leaf node containing the answer text.
In this work, we have opted to omit answer-less questions and question-less answers, however, TreeForm can be extended to included these.

\paragraph{Tables} Tables can be viewed as a collection of entries where each entry has a potential header, and the same set of question-answer pairs.
We thus represent tables through nested trees.
Each entry to the table forms its own node, and contains a leaf node with the entry header or table header if either exist.
We then connect each question-answer pair of the entry as described above.
TreeForm considers tables to be read in row-major order, i.e., we consider each row to be an entry to the table with a potential entry header (row name), and each column to be a specific question that an entry answers.

\paragraph{Free-form Text}
Much like recent work that ignores the \emph{other} entity of FUNSD, we believe that free-form text does not capture the important structural information of a form, and so we do not include such components in TreeForm.
Nevertheless, they can be included as leaf nodes connected to non-leaf nodes (e.g., headers) that they relate to.

\subsection{Concise TreeForm}\label{sec:treeform:concise}
The above structure contains additional information to enable structured prediction and evaluation.
However, it is possible to push textual information up the tree to construct a much denser structure.
We can replace intermediate header and answer nodes by their textual information to reduce both the depth and breadth of the tree.
This can also be done for table entry names (if they exist).
Similarly, we can condense question-answer pairs into a question node whose value is the question text with a child answer node whose value is the answer text.
This concise TreeForm structure removes any unnecessary labeling; for a well structured and well annotated form, this would likely mean all node values will contain the content of the form.
We provide an example of the concise TreeForm structure in \cref{fig:example:concise-treeform}.

\section{Representation of Different Annotation Schemes}\label{app:json}
\cref{fig:example} provides examples of \json annotations for FUNSD, aggregated FUNSD (used for NAA and tree F1 metric), non-concise TreeForm, and concise TreeForm annotations.

\section{Experimental Set-up}\label{app:experiments}
In this section, we discuss the necessary pre- and post-processing steps we took to fine-tune evaluate LayoutXLM and Donut for TreeForm.

\subsection{LayoutXLM} 
LayoutXLM \cite{xu-22-xfund} is a multilingual layout language model for document understanding.
It is an extension of the popular LayoutLM models \cite{xu-21-layout, xu-21-layoutlm2} that were pre-trained for English documents only.
The model takes as input the document image as well tokenized OCR output such that it knows the ground-truth text values in the form. 
The fine-tuned model first predicts the entities which are then used to predict the entity links such that it provides FUNSD-like predictions.\footnote{We fine-tuned the model using the recommended commands and configurations given by \cite{xu-22-xfund} at \url{https://github.com/microsoft/unilm/tree/master/layoutxlm}. Due to resource availability, we use the base version of the model and only trained LayoutXLM with a single GPU. Consequently, our models did not reproduce the results of \cite{xu-22-xfund}.}
We can then evaluate the model's NAA and F1 performance as described in \cref{sec:end-to-end}.
We can further apply the same transformations to the predictions described in \cref{sec:treeform:transform} to get the TreeForm predictions.
Since bounding boxes are already known for each token, they can be used to apply the TreeForm transformations previously described.

\subsection{Donut} 
Donut \cite{kim-22-ocr} is an end-to-end (document) image-to-\json model.
It is pre-trained for a variety of VRDU tasks such as document classification, VQA, and document parsing.
Specifically, Donut was shown to do well in complete document parsing for receipts and invoices, and so applying it to forms is a natural progression.
We note that Donut was developed concurrently with a similar end-to-end model, Dessurt \cite{davis-22-end}.
While Dessurt builds on work more closely related to VRFU \cite{davis-19-deep, davis-21-fudge}, we chose to run experiments using Donut as the model is more accessible for custom training purposes.
Furthermore, Donut was pre-trained with additional synthetic data in Chinese, English, Korean, and Japanese while Dessurt was pre-trained purely on English data.
Therefore, Donut is more suited to be fine-tuned for multiple languages than Dessurt.
More recently, Pix2Struct \citep{lee-2023-pix2struct} has been introduced and generally achieves better performance than Donut.
However, much like Dessurt, it was only trained on English data and so we use Donut in this work.

In order to fine-tune the model\footnote{Donut is available at \url{https://github.com/clovaai/donut}.} to predict FUNSD-like annotations, we create \json objects that represent the trees constructed in \cref{sec:end-to-end}.
This does mean that if a node has multiple parents, its text is repeated several times in the \json representation.
To fine-tune Donut for TreeForm annotations, we use the non-concise TreeForm representation as the model expects all textual components to be leaves in the \json tree.
For both annotation schemes, we follow the training configuration of \cite{kim-22-ocr} for the Consolidated Receipt Dataset (CORD) dataset \cite{park-19-cord}, which predicts a full parse tree for receipts.
We make two minor changes to the training-set up.
Firstly, we use a maximum sequence length of $1024$ rather than $768$ due to forms naturally containing more information than receipts.
Secondly, we fine-tune on the transformed FUNSD dataset for $50$ epochs rather than the $30$ epochs used for CORD as our training set of $149$ documents (per dataset) is much smaller than the $800$ of CORD.

Donut is a generative model, and as such its input is not always formatted in line with the tree structure of the ground truth.
Consequently, we applied a few greedy post-processing transformations that either created valid annotations, or discarded non-viable structures.\footnote{E.g., unanswered questions, empty trees, \emph{inter alia}.}
Furthermore, we observed that the model seemed to sometimes suffer from duplicating text and tree components.\footnote{We suspected this may be due to the increased maximum sequence length, but also saw similar issues when using $768$ as in \cite{kim-22-ocr}.}
As such, we applied a further greedy transformation that removed similar looking leaves and their paths if their normalized Levenshtein distance was greater than $0.6$.
When discarding a path, we always kept the path containing the longest text.
We used a similar heuristic to remove repeated long entities (greater than $20$ characters) for the FUNSD-type prediction.
We only applied the heuristic for long entities as we expect more repetition as previously described.

\end{document}